\documentclass[submission,copyright,creativecommons]{eptcs}
\providecommand{\event}{ICLP 2019} 
\usepackage{breakurl}             
\usepackage{underscore}           

\usepackage{amssymb}
\usepackage{amsthm}
\usepackage{amsmath}
\usepackage{comment}
\usepackage{algorithm}
\usepackage{algorithmic}
\usepackage{moresize}
\usepackage{helvet}
\usepackage{courier}
\usepackage[pdftex]{graphicx}
\usepackage{wrapfig}
\usepackage[font=small]{caption}
\usepackage{subcaption}
\def\i#1{\hbox{\it #1\/}}
\def\beq{\begin{equation}}
\def\eeq#1{\label{#1}\end{equation}}
\def\ba{\begin{array}}
\def\ea{\end{array}}

\def\t{\textbf{t}}
\def\f{\textbf{f}}

\def\iif{\hbox{\bf if}}

\def\causes{\hbox{\bf causes}}

\usepackage[finalold]{trackchanges}

\addeditor{dm}

\addeditor{lb}

\title{A Human-Centered Data-Driven Planner-Actor-Critic Architecture via Logic Programming}
\author{Daoming Lyu
\institute{Auburn University, Auburn, AL, USA}
\email{daoming.lyu@auburn.edu}
\and
Fangkai Yang
\institute{NVIDIA Corporation, Redmond, WA, USA}
\email{fangkaiy@nvidia.com}
\and
Bo Liu \thanks{Correspondence to: Bo Liu$<$\texttt{boliu@auburn.edu}$>$.}
\institute{Auburn University, Auburn, AL, USA}
\email{boliu@auburn.edu}
\and
Steven Gustafson
\institute{Maana Inc., Bellevue, WA, USA}
\email{steven.gustafson@gmail.com}
}

\def\titlerunning{A Human-Centered Data-Driven Planner-Actor-Critic Architecture via Logic Programming}
\def\authorrunning{D. Lyu, F. Yang, B. Liu, \& S. Gustafson}

\begin{document}
\maketitle

\begin{abstract}
Recent successes of Reinforcement Learning (RL) allow an agent to learn policies that surpass human experts but suffers from being time-hungry and data-hungry. By contrast, human learning is significantly faster because prior and general knowledge and multiple information resources are utilized. In this paper, we propose a \textbf{P}lanner-\textbf{A}ctor-\textbf{C}ritic architecture for hu\textbf{MAN}-centered planning and learning (\textbf{PACMAN}), where an agent uses its prior, high-level, deterministic symbolic knowledge to plan for goal-directed actions, and also integrates the Actor-Critic algorithm of RL to fine-tune its behavior towards both environmental rewards and human feedback. This work is the first unified framework where knowledge-based planning, RL, and human teaching jointly contribute to the policy learning of an agent. Our experiments demonstrate that PACMAN leads to a significant jump-start at the early stage of learning, converges rapidly and with small variance, and is robust to inconsistent, infrequent, and misleading feedback.
\end{abstract}


\section{Introduction}

Programming agent that behaves intelligently in a complex domain is one of the central problems of artificial intelligence.  Many tasks that we expect the agent to accomplish can be seen as a ``sequential decision-making'' problem, i.e.,
 the agent makes a series of decisions on how to act in the environment based on its current situation. Recently, reinforcement learning (RL) algorithms have achieved tremendous success involving high-dimensional sensory inputs such as a training agent to play Atari games from raw pixel images~\cite{dqn:nature:2015}.
This approach can learn fine-granular policies that surpass human experts but is criticized for being ``data-hungry" and ``time-hungry". It usually requires several millions of samples. Besides,  it usually has a slow initial learning curve, with a bad performance level of the initial policy before a satisfactory policy can be obtained. By contrast, human learn to play video games significantly faster than the state-of-the-art RL algorithms due to two reasons. First, humans are embodied with prior, general, and abstract knowledge that can be applied and tailored towards a wide class of problems, such that individual problems are treated as special cases \cite{tsividis2017human}. Second, humans learn policy from multiple information resources, including environmental reward signals, human feedback, and demonstrations. 

Theoretical studies that try to simulate human problem-solving from the two aspects above have been done in both knowledge representation(KR) and RL communities. From the first perspective, research from KR community on modular action languages \cite{lif06,erdo08,inclezan2016modular} proposed formal languages to encode a general-purpose library of actions that can be used to define a wide range of benchmark planning problems as special cases, leading to a representation that is elaboration tolerant and addressing the problem of {\em generality of AI} \cite{mcc87}. Meanwhile, researchers from the RL community focused on incorporating high-level abstraction into flat RL, leading to options framework for hierarchical RL \cite{barto-sm:hrl}, hierarchical abstract machines \cite{parr1998reinforcement}, and more recently, works that integrate symbolic knowledge represented in answer set programming (ASP) into reinforcement learning framework \cite{leonetti2016synthesis,yang:peorl:2018,lyu2018sdrl,yang:iros:2019}. From the second perspective, imitation learning, including learning from demonstration (LfD) \cite{lfd:argall:2009} and inverse reinforcement learning (IRL) \cite{ng2000algorithms} tried to learn policies from examples of a human expert, or learn directly from human feedback \cite{thomaz2006reinforcement,knox2010combining,griffith2013policy}, a.k.a, human-centered reinforcement learning (HCRL). In particular, recent studies showed that human feedback should be interpreted in terms of an advantage estimate (a value roughly corresponding to how much better or worse an action is compared to the current
policy) as well to combat positive reward cycles and forgetting, leading to COACH framework \cite{macglashan2016convergent,macglashan2017interactive} based on Actor-Critic (AC) algorithm of RL \cite{nac:Bhatnagar09}. While all of the existing research has shed lights on the importance of prior knowledge and learning from multiple information resources, there is no unified framework to bring them together.

In this paper, we argue that prior knowledge, learning from environmental reward and human teaching can jointly contribute to obtaining the optimal behavior. Prior, explicitly encoded knowledge is general and not sufficient to generate an optimal plan in a dynamic and changing environment, but can be used as a useful guideline to act and learn, leading to a jump start at early stages of learning. The agent further learns domain details to refine its behavior simultaneously from both environmental rewards and human feedback. Thus the prior knowledge is enriched with experience and tailored towards individual problem instances. Based on the motivation above, we propose the \textbf{P}lanner--\textbf{A}ctor--\textbf{C}ritic architecture for hu\textbf{MAN} centered planning and RL (\textbf{PACMAN}) framework. PACMAN interprets human feedback as the advantage function estimation similar to COACH framework and further incorporates prior, symbolic knowledge. The contribution of this paper is as follows.
\begin{itemize}
\item First we propose the \textbf{P}lanner--\textbf{A}ctor--\textbf{C}ritic (PAC) architecture for hu\textbf{MAN} centered planning and RL (\textbf{PACMAN}), featuring symbolic planner-actor-critic trio iteration, where planning and RL mutually benefit each other. In particular, the logical representation of action effects is dynamically generated by sampling a stochastic policy learned from Actor-Critic algorithms of RL. PACMAN allows the symbolic knowledge and actor-critic framework to integrate into a unified framework seamlessly.
\item Second, we enable learning simultaneously from both environmental reward and human feedback, which can accelerate the learning process of interactive RL, and improve the tolerance of misleading feedback from human users. 
\end{itemize}
To the best of our knowledge, this paper is the first work in which learning happens simultaneously from human feedback, environmental reward, and prior symbolic knowledge. While our framework is generic in nature, we choose to use ASP-based action language $\mathcal{BC}$ \cite{lee13}, answer set solver {\sc clingo} to do symbolic planning and conduct our experiment. In principle, each component can be implemented using different techniques.  The evaluation of the framework is performed on RL benchmark problems such as Four Rooms and Taxi domains. Various scenarios of human feedback are considered, including cases of ideal, infrequent, inconsistent, and both infrequent and inconsistent with helpful feedback and misleading feedback. Our experiments show that PACMAN leads to a significant jump-start at early stages of learning, converges faster and with smaller variance, and is robust in inconsistent, infrequent cases even with misleading feedback.

The rest of the paper is organized as follows: Section~\ref{sec:relatedwork} introduces the related work. Section~\ref{sec:prelim} briefly introduces the background about symbolic planning and actor-critic (AC) framework. 
PACMAN architecture is presented in Section~\ref{sec:pacman-arch} and experimental results are shown in Section~\ref{sec:exp-and-res}.

\section{Related Work}
\label{sec:relatedwork}
There is a long history of work that combines symbolic planning with reinforcement learning \cite{parr1998reinforcement,Ryan98rl-tops:an,Ryan02usingabstract,hogg2010learning,leonetti2012automatic,leonetti2016synthesis}. These approaches were based on integrating symbolic planning with value iteration methods of reinforcement learning, and in their work, there was no bidirectional communication loop between planning and learning so that they could not mutually benefit each other. The latest work in this direction is PEORL framework \cite{yang:peorl:2018} and SDRL \cite{lyu2018sdrl}, where ASP-based planning was integrated with R-learning \cite{rlearning:schwartz} into planning--learning loop. PACMAN architecture is a new framework of integrating symbolic planning with RL, in particular, integrating planning with AC algorithm for the first time, and also features bidirectional communication between planning and learning.

Learning from human feedback takes the framework of reinforcement learning, and incorporate human feedback into the reward structure \cite{thomaz2006reinforcement,knox2009interactively,knox2012reinforcement}, information directly on policy \cite{thomaz2008teachable,knox2010combining,griffith2013policy}, or advantage function \cite{macglashan2016convergent,macglashan2017interactive}.
Learning from both human feedback and environmental rewards were investigated \cite{thomaz2006reinforcement,knox2012reinforcement,griffith2013policy}, mainly integrating the human feedback to reward or value function via reward shaping or Q-value shaping. 
Such methods do not handle well the samples with missing human feedback, and in reality, human feedback may be infrequent. They also suffer from the ambiguity of statement-like reward such as ``that's right'' or ``no, this is wrong'': such statements are easy to be transformed to binary or discrete value but are difficult to incorporate with continuous-valued reward and value signals, as pointed out by \cite{griffith2013policy}.
Recent work of COACH \cite{macglashan2016convergent,macglashan2017interactive} showed that human feedback was more likely to be policy-dependent, and advantage function provides a better model of human feedback, but it does not consider learning simultaneously from environmental reward and human feedback. Furthermore, none of these work considers the setting where an agent is equipped with prior knowledge and generates a goal-directed plan that is further to be fine-tuned by reinforcement learning and a human user. Integrating COACH-style human feedback into PACMAN, our framework allows the integration of symbolic planning into the learning process, where environmental reward and human feedback can be unified into the advantage function to shape the agent's behavior in the context of long-term planning. 

\section{Preliminaries}
\label{sec:prelim}
This section introduces the basics of symbolic planning and actor-critic framework.

\subsection{Symbolic Planning}
Symbolic planning is concerned with describing preconditions and effects of actions using a formal language and automating plan generation. An {\em action description} $D$ in the language $\mathcal{BC}$ includes two kinds of symbols, {\em fluent constants} that represent the properties of the world, with the signature denoted as $\sigma_F(D)$, and {\em action constants}, with the signature denoted as $\sigma_A(D)$. A \textit{fluent atom} is an expression of the form $f=v$, where $f$ is a fluent constant and $v$ is an element of its domain. For the Boolean domain, denote $f=\t$ as $f$ and~$f=\f$ as $\sim\!\!\! f$. An action description is a finite set of {\em causal laws} that describe how fluent atoms are related with each other in a single time step, or how their values are changed from one step to another, possibly by executing actions. For instance,
\beq
A~\iif~A_1,\ldots,A_m
\eeq{static}
is a {\em static law} that states at a time step, if $A_1,\ldots, A_m$ holds then $A$ is true. 
\beq
a~\causes~A_0~\iif~A_1,\ldots, A_m
\eeq{dynamic}
is a {\em dynamic law}, stating that at any time step, if $A_1,\ldots, A_m$ holds, by executing action $a$,~$A_0$ holds in the next step.\footnote{In $\mathcal{BC}$, causal laws are defined in a more general form. In this paper, without loss of generality, we assume the above form of causal laws for defining effects of actions.} An action description captures a dynamic transition system. 
Let $I$ and $G$ be states. The triple $(I,G,D)$ is called a planning problem. $(I,G,D)$ has a plan of length $l-1$ iff there exists a transition path of length $l$ such that $I=s_1$ and $G=s_l$. Throughout the paper, we use $\Pi$ to denote both the plan and the transition path by following the plan. Genearating a plan of length $l$ can be achieved by solving the answer set program $\i{PN}_l(D)$, consisting of rules translated from $D$ and appending timestamps from 1 to $l$, via a translating function $\i{PN}$. For instance, $\i{PN}_l$ turns (\ref{static}) to
$$
i\!:\!A\leftarrow i\!:\!A_1,\ldots, i\!:\!A_m,
$$
where $1\le i\le l$
and (\ref{dynamic}) to 
$$
i+1\!:\!A\leftarrow i\!:\!a, i\!:\!A_1,\ldots, i\!:\!A_m,
$$
where $1\le i < l$. See \cite{lee13} for details.

\subsection{Actor-Critic Architecture}

\begin{figure}[htb!]
\centering
\includegraphics[height=5.2cm,width=5.3cm]{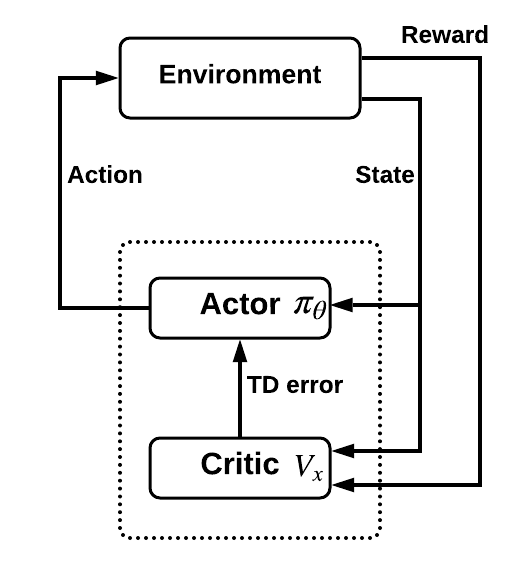}
\caption{The framework of actor-critic}
\label{fig:actor-critic}
\end{figure}

A Markov Decision Process (MDP) is defined as a tuple $({\mathcal{S},\mathcal{A},P_{ss'}^{a},r,\gamma})$, where $\mathcal{S}$ and $\mathcal{A}$ are the sets of symbols denoting state space and action space, the transition kernel $P_{ss'}^{a}$ specifies the probability of transition from state $s\in\mathcal{S}$ to state $s'\in\mathcal{S}$ by taking action $a\in\mathcal{A}$, $r(s,a):\mathcal{S}\times\mathcal{A}\mapsto\mathbb{R}$ is a reward function bounded by $r_{\max}$, and $0\leq\gamma<1$ is a discount factor. A solution to an MDP is a policy $\pi:\mathcal{S}\mapsto \mathcal{A}$ that maps a state to an action. RL concerns learning a near-optimal policy by executing actions and observing the state transitions and rewards, and it can be applied even when the underlying MDP is not explicitly given.

An actor-critic~\cite{nac:peters2008,nac:Bhatnagar09} approach is a framework of reinforcement learning, which has two components: the actor and the critic, as shown in Figure~\ref{fig:actor-critic}. Typically, the actor is a policy function $\pi_{\theta}$ parameterized by $\theta$ for action selection, while the critic is a state-value function $V_{x}$ parameterized by $x$ to criticize the action made by the actor. For example, after each action selection, the critic will evaluate the new state to determine whether things have gone better or worse than expected by computing the temporal difference (TD) error \cite{sutton2018reinforcement}, $$\delta(s,a,s') = r(s,a)+\gamma V_x(s')- V_x(s).$$ 
If the TD error is positive, it suggests that the tendency to select current action $a$ should be strengthened for the future, whereas if the TD error is negative, it suggests the tendency should be weakened. This TD error is actually an estimate of advantage function \cite{schulman2015high}.



\section{PACMAN Architecture}
\label{sec:pacman-arch}

In this section, we will present our PACMAN in detail. 

\subsection{Sample-based Symbolic Planning}
\begin{figure*}[!tbp]
\centering
\includegraphics[width=0.99\textwidth]{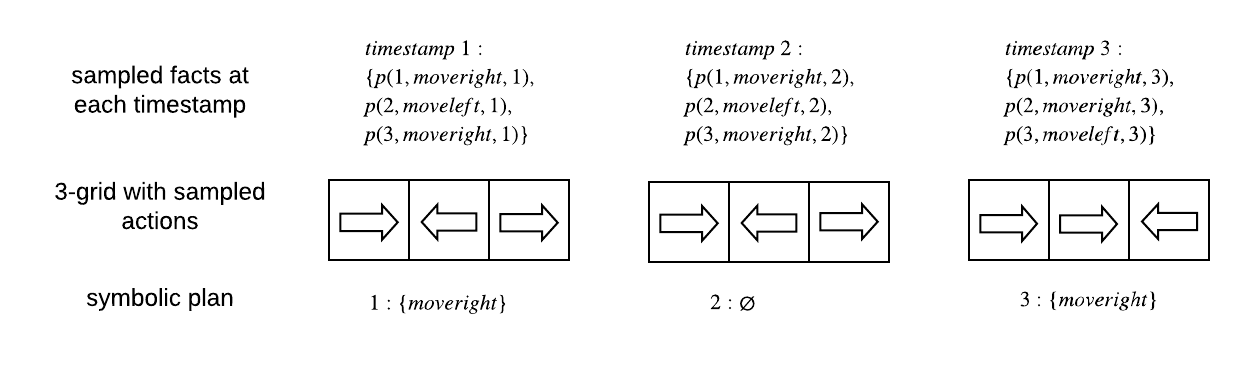}
\caption{A possible sample-based planning result for 3-grid domain}
\label{fig:sample}
\end{figure*}
We introduce a {\em sample-based planning problem} as a tuple $(I,G,D,\pi_{\theta})$ where $I$ is the initial state condition, $G$ is a goal state condition, $D$ is an action description in $\mathcal{BC}$, and~$\pi_{\theta}$ is a stochastic policy function parameterized by~$\theta$, i.e., a mapping $\mathcal{S}\times \mathcal{A} \mapsto [0,1]$. For $D$, defines its $l$-step {\em sampled action description} $D^l_{\pi} = D_s\cup D_d\cup \bigcup_{t=1}^l P^t_{\pi}$ with respect to policy $\pi$ and time stamp $1\le t\le l$, where
\begin{itemize}
\item $D_s$ is a set of causal laws consisting of static laws and dynamic laws that does not contains action symbols;
\item $D_d$ is a set of causal laws obtained by turning each dynamic law of the form
$$
a~\causes~A_0~\iif~A_1,\ldots,A_m,
$$
into rules of the form
$$
a~\causes~A_0~\iif~A_1,\ldots,A_m,p(s,a)
$$
where $\i{p}$ is a newly introduced fluent symbol and $\{A_1,\ldots, A_m\}\subseteq s$, for $s\in\mathcal{S}$; and
\item $P^t_{\pi}$ is a set of facts sampled at timestamp $t$ that contains $p(a,s)$ such that
$$
p(s,a)\in P^t_{\pi}\sim \pi_{\theta}(\cdot|s)
$$
where for $s\in\mathcal{S}$, $A\in\mathcal{A}$.
\end{itemize}
Define the translation $\mathcal{T}(D^l_{\pi})$ as
$$
\i{PN}_l(D_s\cup D_d)\cup\bigcup_{t=1}^l\{p(s,a,t), \hbox{for}~p(s,a)\in P^t_\pi\}
$$
that turns $D^l_\pi$ into answer set program.

A {\em sample-based plan} up to length $l$ of $(I,G,D,\pi_{\theta})$ can be calculated from the answer set of program $\mathcal{T}(D^l_{\pi})$ such that $I$ and $G$ are satisfied. The planning algorithm is shown in Algorithm~\ref{alg:splanning}.

\begin{algorithm}[!tbp]
  \caption{Sample-based Symbolic Planning}
  \label{alg:splanning}
  \begin{algorithmic}[1]
    \REQUIRE a sample based planning problem $(I,G,D,\pi_{\theta})$
    \STATE $\Pi\Leftarrow\emptyset$ 
    \STATE calculate $D_{\pi}^0$ 
    \STATE $k\Leftarrow 1$
      \WHILE{$\Pi=\emptyset$ and $k<\i{maxstamp}$}    
        \STATE sample $P_{\pi,k}$ over $p(a,s)\sim\pi_{\theta}(\cdot|s)$ for $s\in\mathcal{S}$, $a\in\mathcal{A}$
        \STATE $D^k_{\pi} \Leftarrow D^{k-1}_{\pi}\cup P^k_{\pi}$
        \STATE $\displaystyle\Pi\leftarrow\textsc{Clingo}.\i{solve}(I\cup G\cup {\mathcal{T}}(D^k_{\pi}))$
        \STATE $k\leftarrow k+1$
      \ENDWHILE
   \RETURN $\Pi$
  \end{algorithmic}
\end{algorithm}

\noindent\textbf{Example.} Consider 3$\times$1 horizontal gridworld where the grids are marked as state 1, 2, 3, horizontally. Initially the agent is located in state 1. The goal is to be located in state 3. The agent can move to left or right. Using action language ${\mathcal{BC}}$, moving to the left and moving to the right can be formulated as dynamic laws
$$
\ba{rl}
\!\!\!\i{moveleft}&\!\!\!\causes~\i{Loc}=L-1~\iif~\i{Loc}=L.\\
\!\!\!\i{moveright}&\!\!\!\causes~\i{Loc}=L+1~\iif~\i{Loc}=L.
\ea
$$
Turning them into sample-based action description leads to
$$
\ba{rlr}
\i{moveleft}&\!\!\!\causes~\i{Loc}=L-1~\iif~\i{Loc}=L,p(L,\i{moveleft}).\\
\i{moveright}&\!\!\!\causes~\i{Loc}=L+1~\iif~\i{Loc}=L,p(L,\i{moveright}).
\ea
$$
A policy estimator $\pi_{\theta}$ accepts an input state and output probability distribution on actions $\i{moveleft}$ and $\i{moveright}$. Sampling $\pi_{\theta}$ with input $s$ at time stamp $i$ generates a fact of the form $p(a,s,i)$ where $a\in\{\i{moveleft},\i{moveright}\}$ following the probability distribution of $\pi_{\theta}(\cdot|s)$.

At any timestamp, \textsc{clingo} solves answer set program consisting of rules translated from the above causal laws:
\begin{verbatim}
    loc(L-1,k+1):-moveleft(k),loc(L,k), p(L,moveleft,k).
    loc(L+1,k+1):-moveright(k),loc(L,k), p(L,moveright,k).
\end{verbatim}
\noindent for time stamp $1,\ldots,k$, plus a set of facts of the form {\tt p(a,s,i)} sampled from $\pi_{\theta}$ where for states $s\in\{1,2,3\}$ and timestamps $i\in\{1,\ldots,k\}$. Note that the planner can skip time stamps if there is no possible actions to use to generate plan, based on sampled results. Figure~\ref{fig:sample} shows a possible sampling results over 3 timestamps, and a plan of 2 steps is generated to achieve the goal, where time stamp 2 is skipped with no planned actions.

Since sample-based planning calls a policy approximator as an oracle to obtain probability distribution and samples the distribution to obtain available actions, it can be easily applied to other planning techniques such as PDDL planning. For instance, the policy appropriator can be used along with heuristics on relaxed planning graph \cite{helmert2006fast}.

\subsection{Planning and Learning Loop}

The planning and learning loop for PACMAN, as shown in Algorithm~\ref{alg:pac}, starts from a random policy (uniform distribution over action space), and then generate a sample-based symbolic plan. After that, it follows the plan to explore and update the policy function $\pi_{\theta}$, leading to an improved policy, which is used to generate the next plan.

\begin{algorithm}
\caption{Planner-Actor-Critic (PAC) Learning}
\label{alg:pac}
\begin{algorithmic}[1]
\REQUIRE $(I,G,D,\pi_{\theta})$ and a value function estimator $V$
\STATE $\i{episode}=0$
\FOR {$\i{episode}<\i{maxepisode}$}
\STATE Generate symbolic plan $\Pi$ from $(I,G,D,\pi_{\theta})$ by Algorithm~\ref{alg:splanning}
\FOR {$\langle s_{i}, a_{i}, r_{i}, s_{i+1} \rangle\in\Pi$}
\STATE Compute TD error as in Eq.~\eqref{eq:tderror} as a stochastic estimation of advantage function.
\STATE Update $V_x$ via Eq.~\eqref{eq:x-td-update}. 
\IF{human feedback $f_i$ is available}
\STATE Replace TD error with human feedback $f_i$.
\ENDIF
\STATE Update $\pi_{\theta}$ via Eq.~\eqref{eq:theta-td-update}. 
\ENDFOR
\STATE $\i{episode}\leftarrow \i{episode}+1$
\ENDFOR
\end{algorithmic}
\end{algorithm}

At the $i$-th time step, the sample is formulated as $( s_{i}, a_{i}, r_{i}, s_{i+1})$, then the TD error is computed as
\begin{align}
{\delta _i} = {r_i} + \gamma {V_{{x_{i+1}}}}({s_{i + 1}}) - {V_{{x_{i+1}}}}({s_i}),
\label{eq:tderror}
\end{align}
which is a stochastic estimation of the advantage function. 
The value function $V_{x}$ is updated using reinforcement learning approaches, such as TD method~\cite{sutton2018reinforcement}:
\begin{align}
{x_{i + 1}} = {x_i} + \alpha {\delta _i}\nabla {V_{{x_i}}}({s_i}),
\label{eq:x-td-update}
\end{align}
where $\alpha$ is the learning rate.
The policy function $\pi_{\theta}$ will be updated by
\begin{align}
\theta_{i+1} = \theta_{i} + \beta {\delta _i}\nabla \log \pi_{\theta_{i}}(a_i|s_i),
\label{eq:theta-td-update}
\end{align}
where $\beta$ is the learning rate.
If the human feedback signal $f_i$ is available, then this feedback signal will replace the previous computed TD error and be used to update the policy function; If there is no human feedback signal available at this iteration, TD error will be used to update the policy function directly. For this reason, human feedback here can be interpreted as guiding exploration towards human preferred state-action pairs.


\section{Experiment}
\label{sec:exp-and-res}

We evaluate our method in two RL-benchmark problems: Four Rooms \cite{sutton1999between} and Taxi domain \cite{barto-sm:hrl}. For all domains, we consider the discrete value of (positive or negative) feedback with the cases of ideal (feedback is always available without reverting), infrequent (only giving feedback at 50\% probability), inconsistent (randomly reverting feedback at 30\% probability) and infrequent+inconsistent (only giving feedback at 50\% probability, while randomly reverting feedback at 30\% probability). We compare the performance of PACMAN with 3 methods: TAMER+RL Reward Shaping from \cite{knox2012reinforcement}, BQL Reward Shaping from \cite{griffith2013policy}, and PACMAN without symbolic planner (AC with Human Feedback) as our ablation analysis. All plotting curves are averaged over 10 runs, and the shadow around the curve denotes the variance.

\subsection{Four Rooms}
\begin{figure*}[htb!]
\begin{subfigure}{.33\textwidth}
\centering
    \includegraphics[height=4.6cm,width=5.0cm]{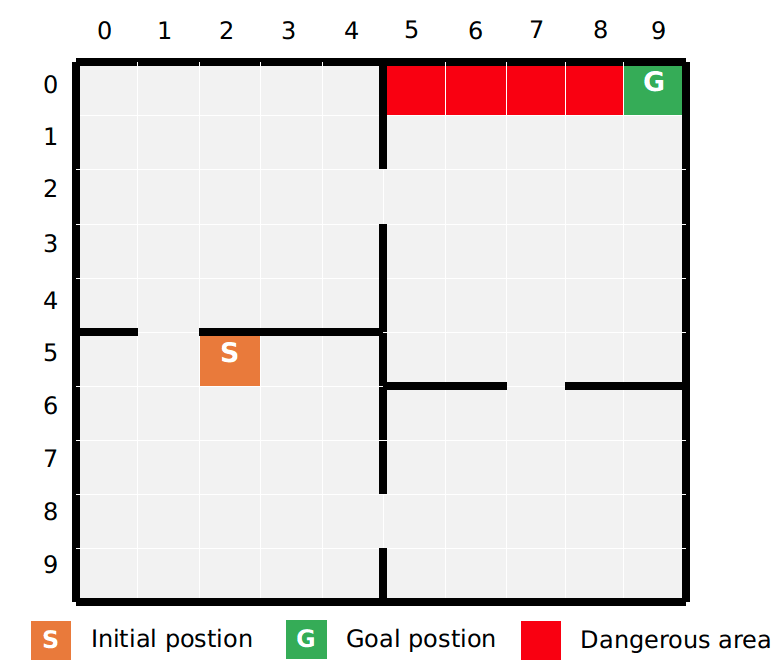}
  \subcaption{Four-room domain}
  \label{fig:4room}
 \end{subfigure}
\begin{subfigure}{.33\textwidth}
\centering
    \includegraphics[height=4.5cm,width=4.5cm]{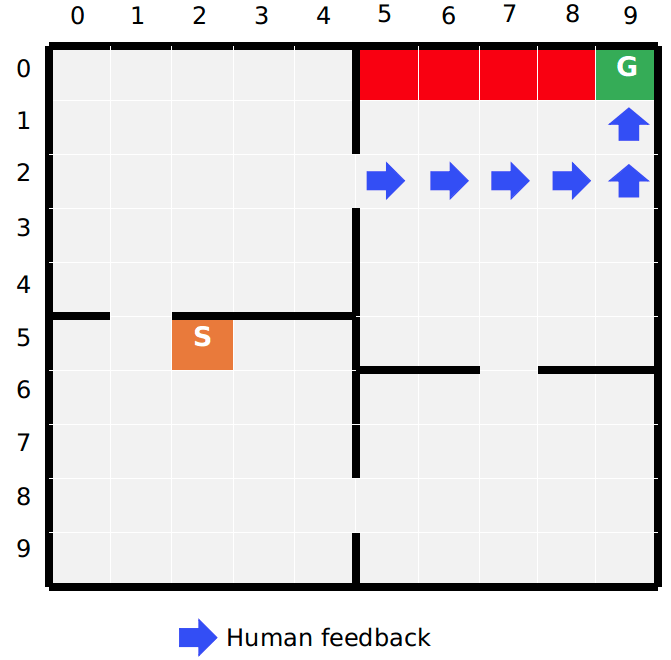}
  \subcaption{Helpful Feedback}
  \label{fig:4room-help}
 \end{subfigure}
 \begin{subfigure}{.33\textwidth}
\centering
    \includegraphics[height=4.5cm,width=4.4cm]{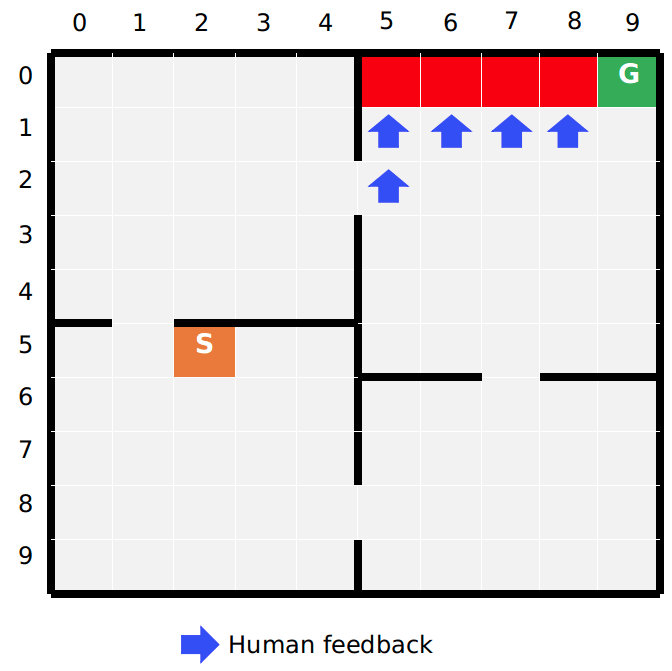}
  \subcaption{Misleading Feedback}
  \label{fig:4room-mislead}
 \end{subfigure}
\caption{The snapshot of 2 scenarios on Four Rooms domain}
\label{fig:4room-2scens}
\end{figure*}

Four Rooms domain is shown in Figure~\ref{fig:4room}. In this 10$\times$10 grid, there are 4 rooms and an agent navigating from the initial position (5,2) to the goal position (0,9). If the agent can successfully achieve the task, it would receive a reward of +5. And it may obtain a reward of -10 if the agent steps into the red grids (dangerous area). Each move will cost -1.

The human feedback of Four Rooms domain concerns 2 scenarios.
\begin{itemize}
\item \textbf{Helpful feedback}. Consider an experienced user that wants to help the agent to navigate safer and better, such that the agent can stay away from the dangerous area and reach the goal position with the shortest path. Therefore, human feedback can guide the agent to improve its behavior towards the task, as shown in Figure~\ref{fig:4room-help}.
\item \textbf{Misleading feedback}. Consider an inexperienced user who doesn't know there is a dangerous area, but wants the agent to step into those red grids (Figure~\ref{fig:4room-mislead}). In this case, human feedback contradicts with the behavior that the agent learns from an environmental reward.
\end{itemize}


\begin{figure*}[htb!]
\begin{subfigure}{.495\textwidth}
    \includegraphics[height=4.1cm,width=6.8cm]{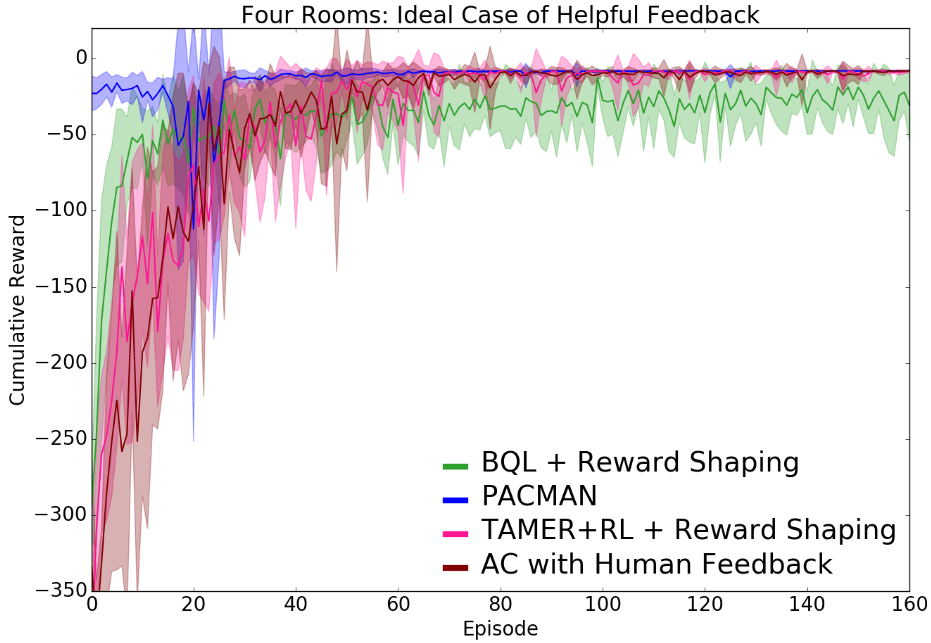}
  \subcaption{Ideal Case}
  \label{fig:ideal-4room-help}
 \end{subfigure}
 \begin{subfigure}{.495\textwidth}
    \includegraphics[height=4.1cm,width=6.8cm]{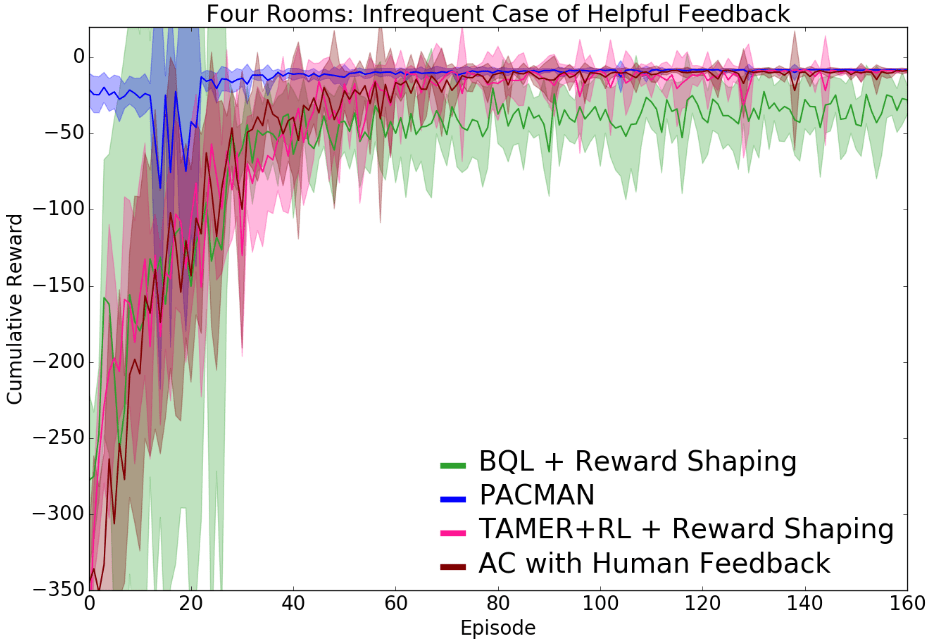}
  \subcaption{Infrequent Case}
  \label{fig:infrequent-4room-help}
\end{subfigure}
 \begin{subfigure}{.495\textwidth}
    \includegraphics[height=4.1cm,width=6.8cm]{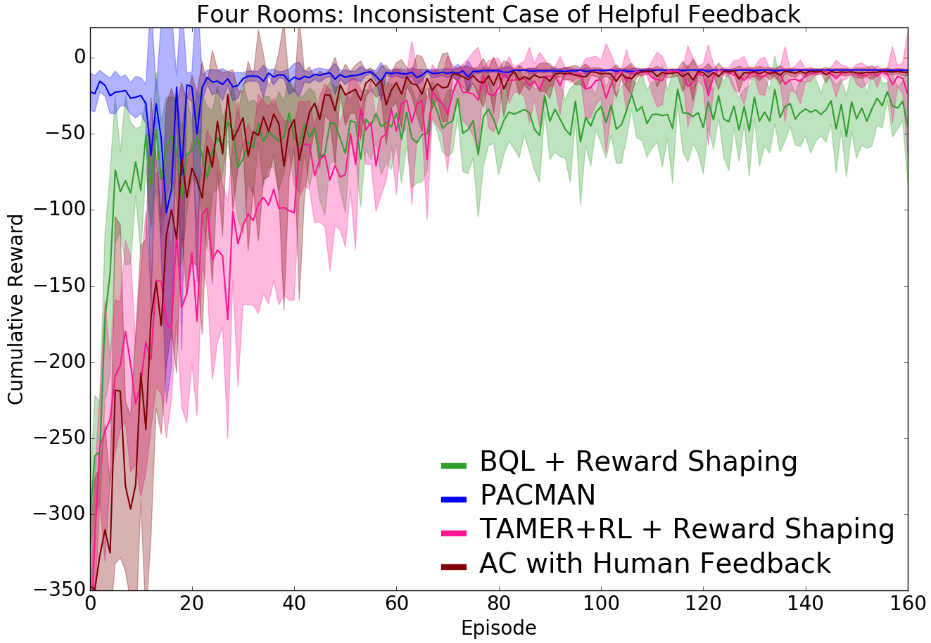}
  \subcaption{Inconsistent Case}
  \label{fig:inconsistent-4room-help}
\end{subfigure}
 \begin{subfigure}{.495\textwidth}
    \includegraphics[height=4.1cm,width=6.8cm]{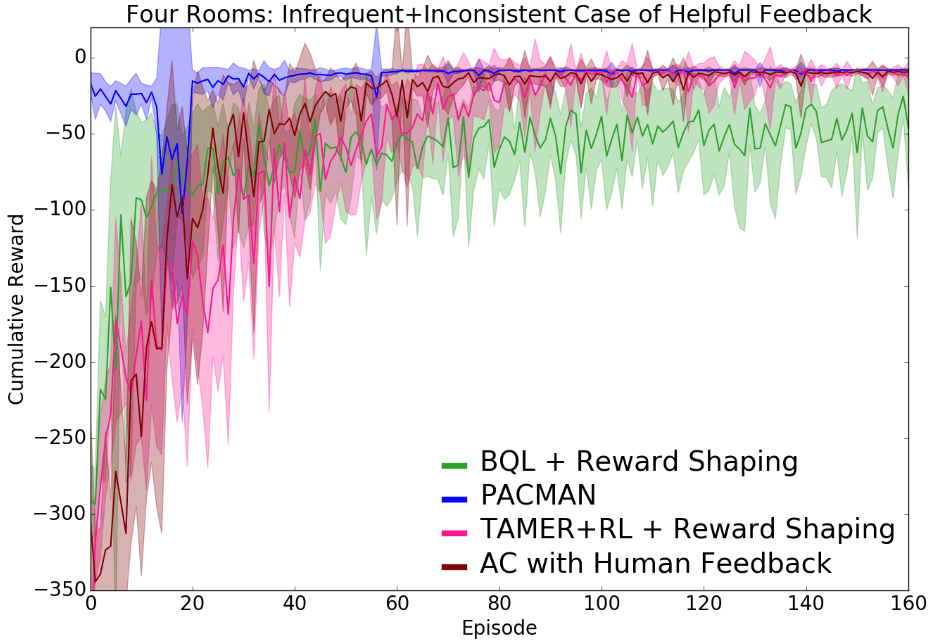}
  \subcaption{Infrequent+Inconsistent Case}
  \label{fig:incons-infreq-4room-help}
\end{subfigure}
\caption{Four-room with Helpful Feedback: Learning Curves}
\label{4roomhelpful}
\end{figure*}

\begin{figure*}[htb!]
\begin{subfigure}{.495\textwidth}
    \includegraphics[height=4.1cm,width=6.8cm]{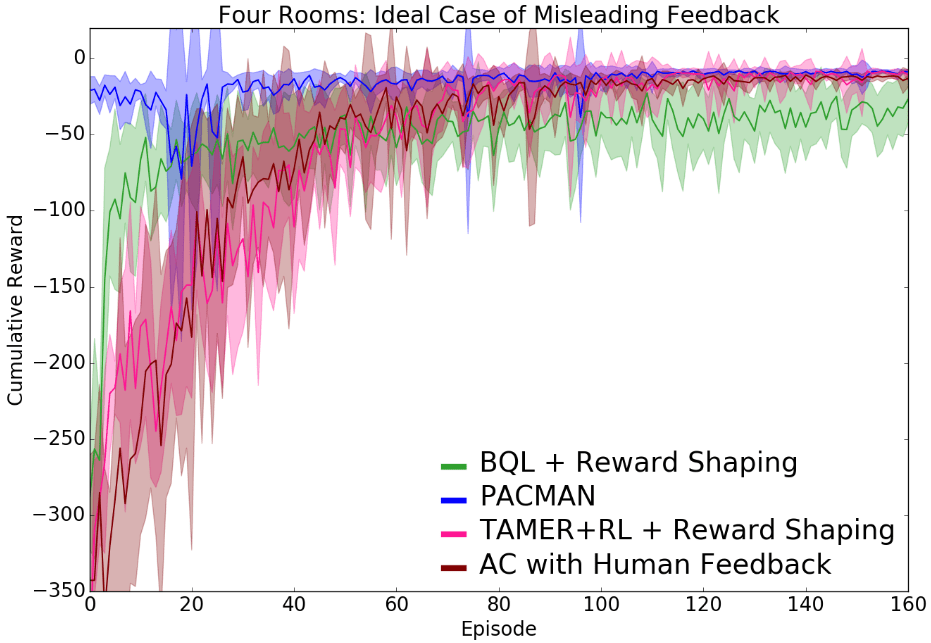}
  \subcaption{Ideal Case}
  \label{fig:ideal-4room-mislead}
 \end{subfigure}
 \begin{subfigure}{.495\textwidth}
    \includegraphics[height=4.1cm,width=6.8cm]{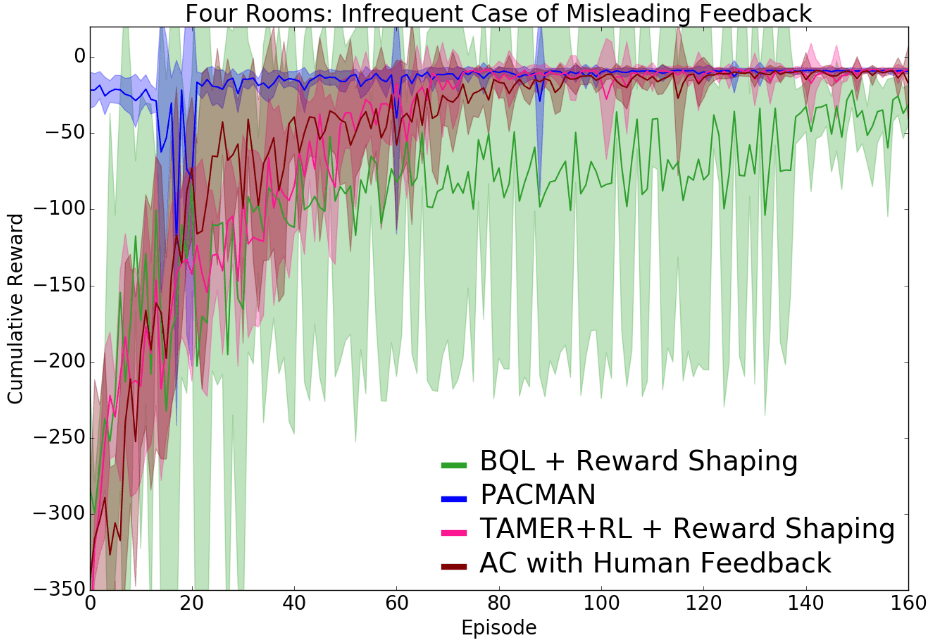}
  \subcaption{Infrequent Case}
  \label{fig:infrequent-4room-mislead}
\end{subfigure}
 \begin{subfigure}{.495\textwidth}
    \includegraphics[height=4.1cm,width=6.8cm]{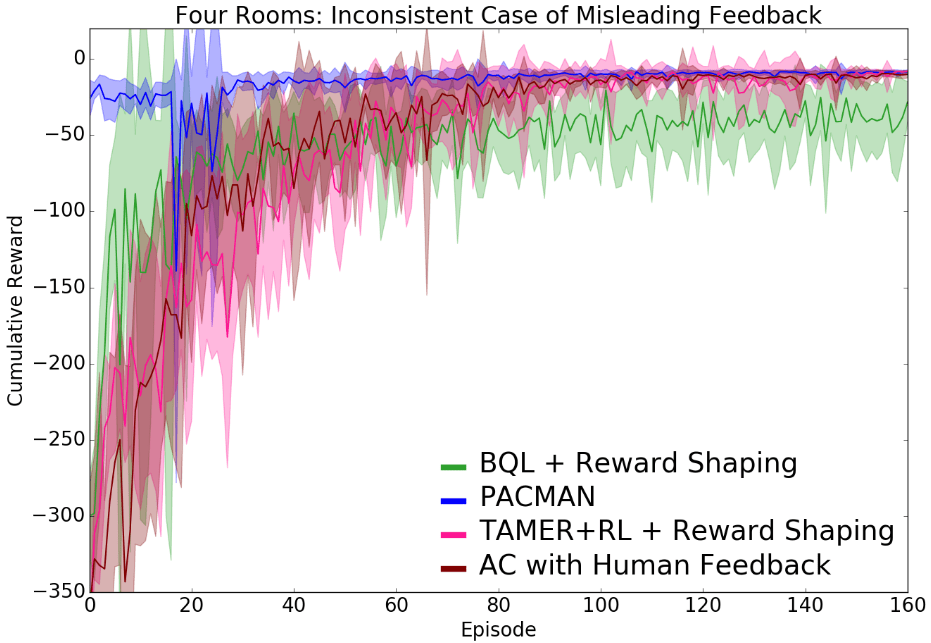}
  \subcaption{Inconsistent Case}
  \label{fig:inconsistent-4room-mislead}
\end{subfigure}
 \begin{subfigure}{.495\textwidth}
    \includegraphics[height=4.1cm,width=6.8cm]{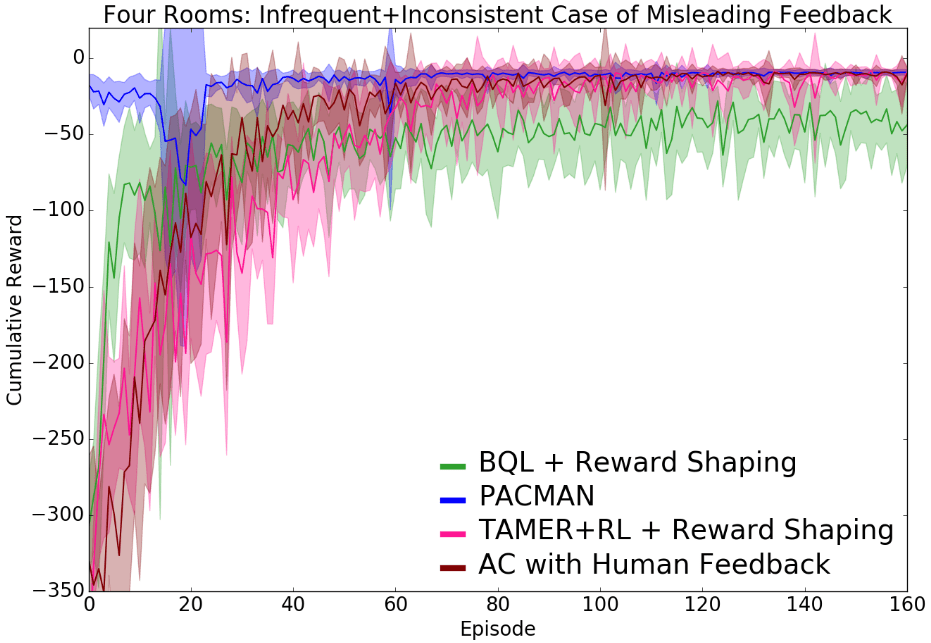}
  \subcaption{Infrequent+Inconsistent Case}
  \label{fig:incons-infreq-4room-mislead}
\end{subfigure}
\caption{Four-room with Misleading Feedback: Learning Curves}
\label{4roommislead}
\end{figure*}


The results are shown in Figure~\ref{4roomhelpful} and Figure~\ref{4roommislead}. As we can see, PACMAN has a jump-start and quickly converged with small variance,
compared to BQL Reward Shaping, TAMER+RL Reward Shaping, and AC with Human Feedback under four different cases. This is because symbolic planning can lead to goal-directed behavior, which would bias exploration. Though the infrequent case, inconsistent case, and their combination case for both helpful feedback and misleading feedback can lead to more uncertainty, the performance of PACMAN remains unaffected, which means more robust than others.
Meticulous readers may find that there is a large variance in the initial stage of PACMAN, especially in Figure~\ref{4roomhelpful} and Figure~\ref{4roommislead}, this is due to the reason that the symbolic planner will first generate a short plan that is reasonably well, then the symbolic planner will perform exploration by generating longer plans. After doing the exploration, the symbolic planner will converge to the short plan with the optimal solution. This large variance can be partially alleviated by setting the maximal number of actions in a plan to reduce plan space.

\subsection{Taxi Domain}
\begin{figure*}[htb!]
 \begin{subfigure}{.33\textwidth}
\centering
    \includegraphics[height=4.9cm,width=4.9cm]{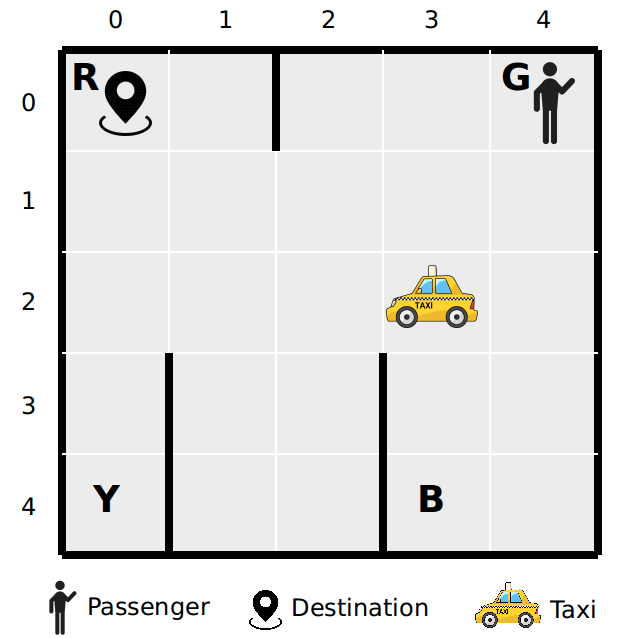}
  \subcaption{Taxi domain}
  \label{fig:taxi}
 \end{subfigure}
\begin{subfigure}{.33\textwidth}
\centering
    \includegraphics[height=4.9cm,width=4.9cm]{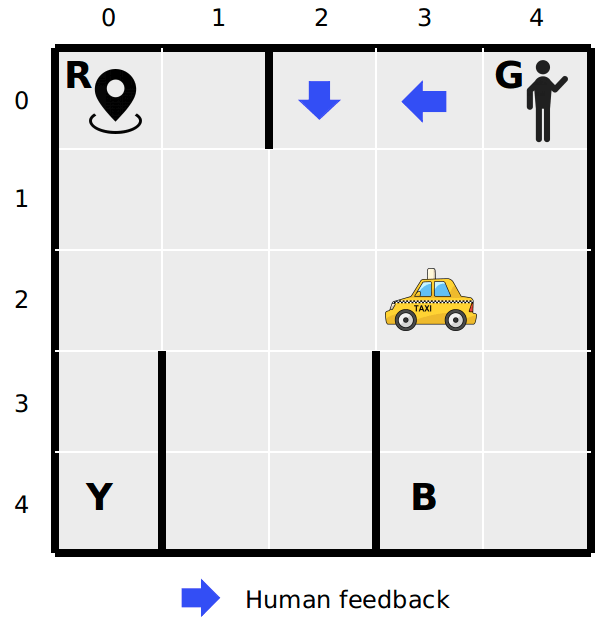}
  \subcaption{Helpful Feedback}
  \label{fig:taxi-help}
 \end{subfigure}
 \begin{subfigure}{.33\textwidth}
\centering
    \includegraphics[height=4.9cm,width=4.7cm]{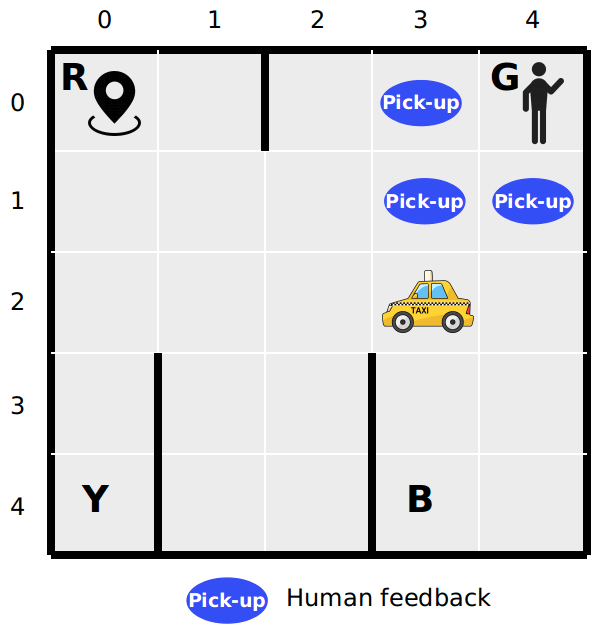}
  \subcaption{Misleading Feedback}
  \label{fig:taxi-mislead}
 \end{subfigure}
\caption{The snapshot of 2 scenarios on Taxi domain}
\label{fig:taxi-2scens}
\end{figure*}

Taxi domain concerns a 5$\times$5 grid (Figure~\ref{fig:taxi}) where a taxi needs to navigate to a passenger, pick up the passenger, then navigate to the destination and drop off the passenger. Each move has a reward of -1. Successful drop-off received a reward of +20, while improper pick-up or drop-off would receive a reward of -10. When formulating the domain symbolically, we specify that precondition of performing picking up a passenger is that the taxi has to be located in the same place as the passenger.

\begin{figure*}[htb!]
\begin{subfigure}{.495\textwidth}
    \includegraphics[height=4.15cm,width=6.8cm]{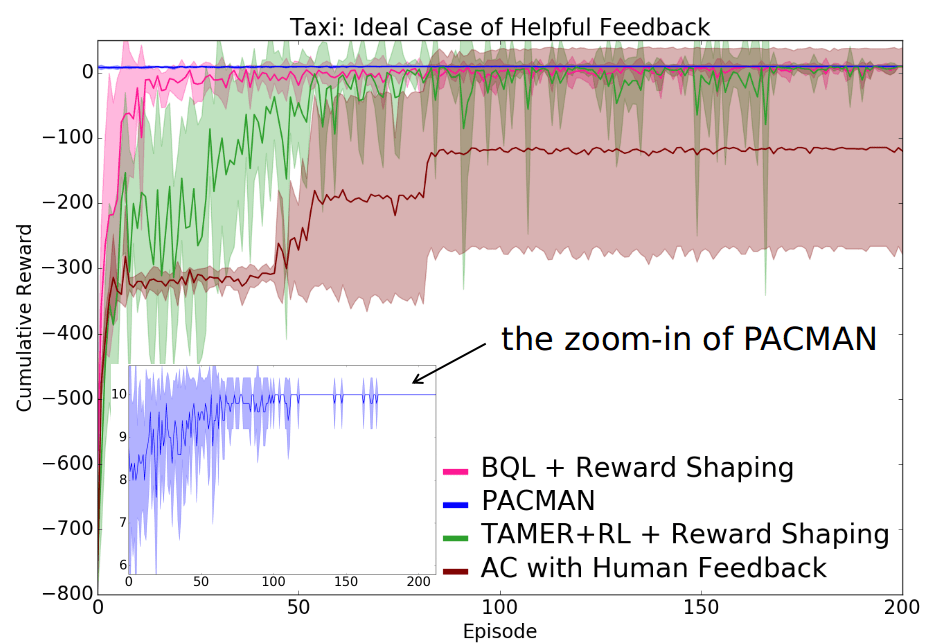}
  \subcaption{Ideal Case}
  \label{fig:ideal-taxi-help}
 \end{subfigure}
 \begin{subfigure}{.495\textwidth}
    \includegraphics[height=4.15cm,width=6.8cm]{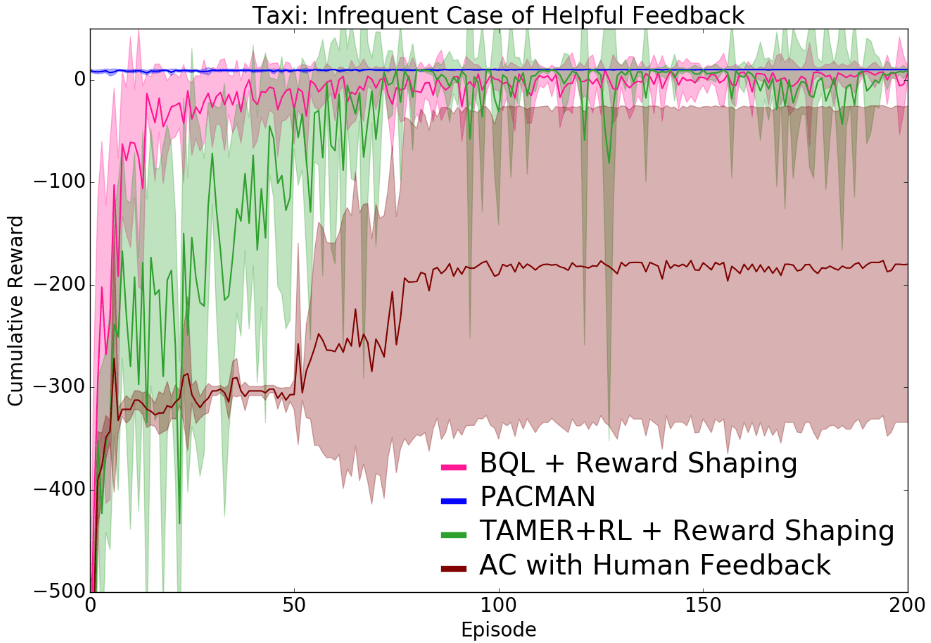}
  \subcaption{Infrequent Case}
  \label{fig:infrequent-taxi-help}
\end{subfigure}
 \begin{subfigure}{.495\textwidth}
    \includegraphics[height=4.15cm,width=6.8cm]{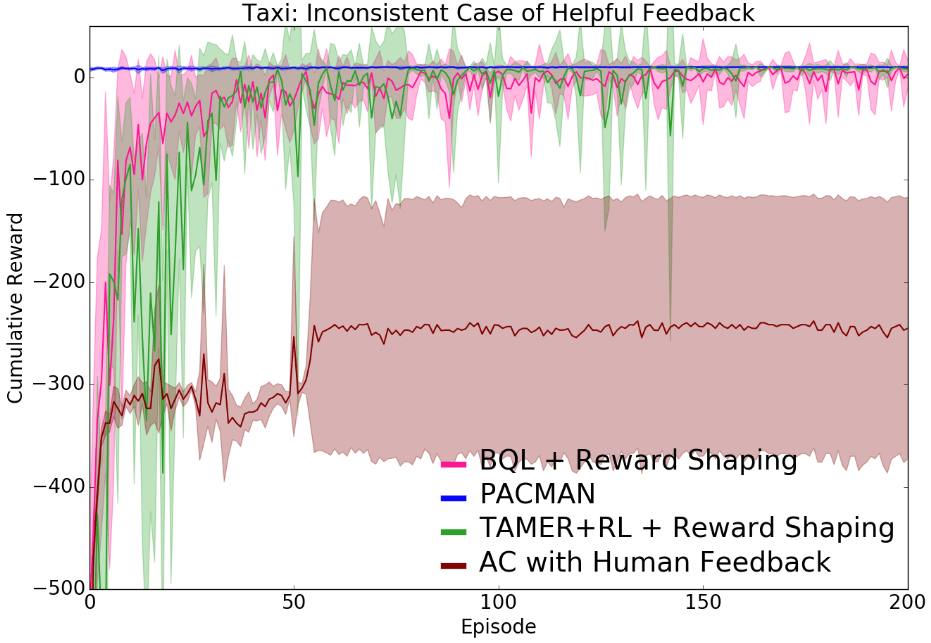}
  \subcaption{Inconsistent Case}
  \label{fig:inconsistent-taxi-help}
\end{subfigure}
 \begin{subfigure}{.495\textwidth}
    \includegraphics[height=4.15cm,width=6.8cm]{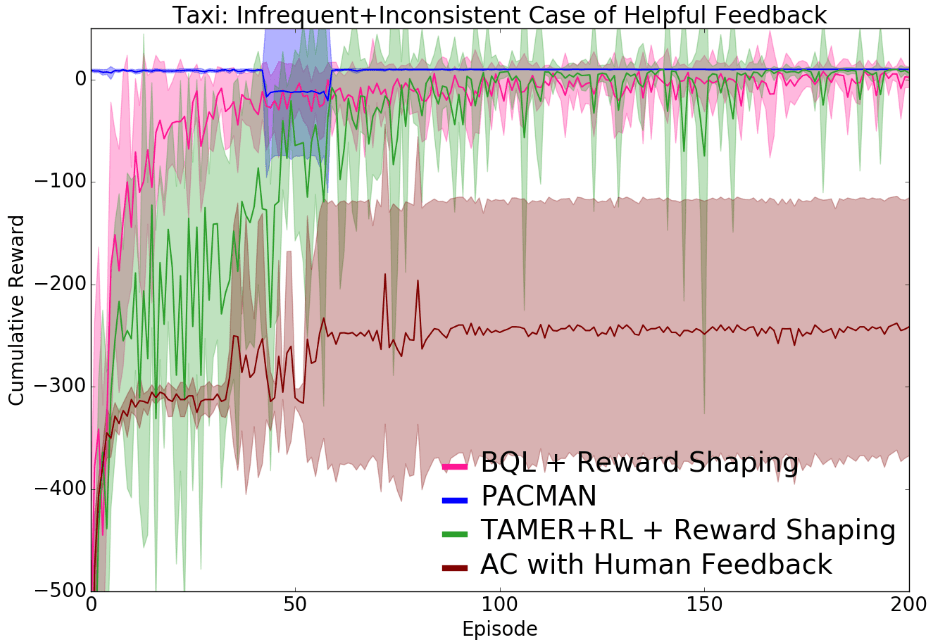}
  \subcaption{Infrequent+Inconsistent Case}
  \label{fig:incons-infreq-taxi-help}
\end{subfigure}
\caption{Taxi with Helpful Feedback: Learning Curves}
\label{taxihelpful}
\end{figure*}

\begin{figure*}[htb!]
\begin{subfigure}{.495\textwidth}
    \includegraphics[height=4.15cm,width=6.8cm]{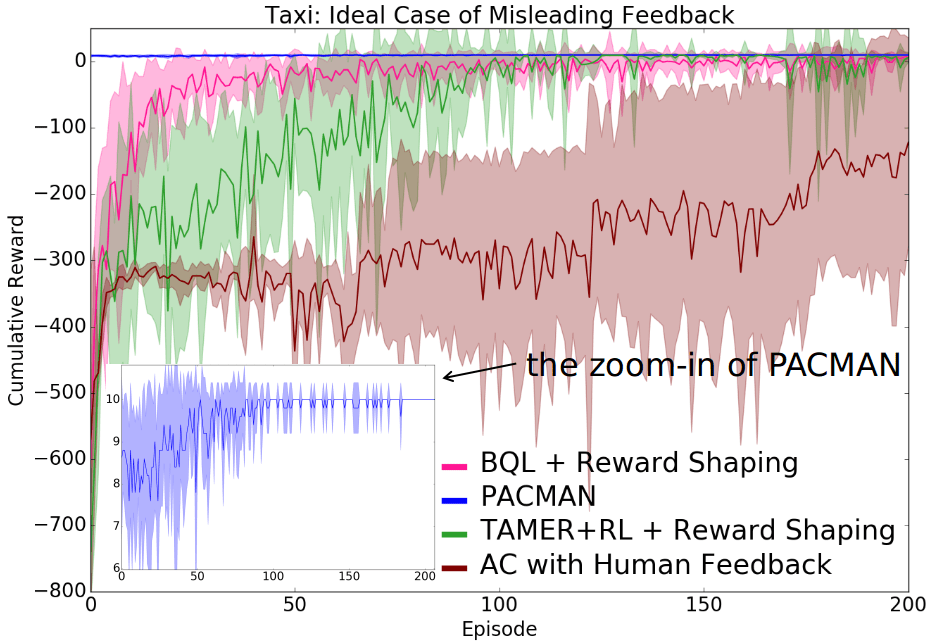}
  \subcaption{Ideal Case}
  \label{fig:ideal-taxi-mislead}
 \end{subfigure}
 \begin{subfigure}{.495\textwidth}
    \includegraphics[height=4.15cm,width=6.8cm]{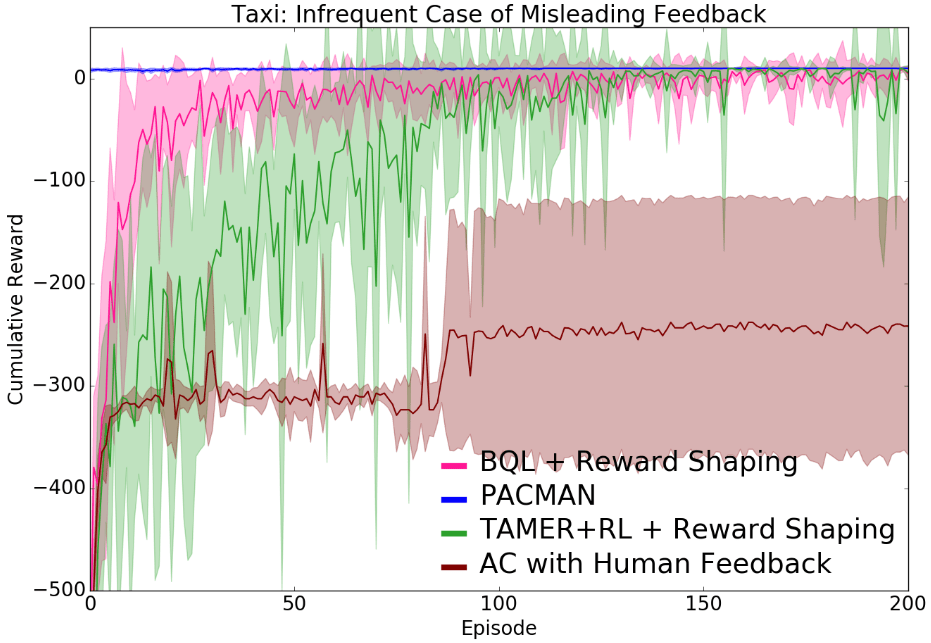}
  \subcaption{Infrequent Case}
  \label{fig:infrequent-taxi-mislead}
\end{subfigure}
 \begin{subfigure}{.495\textwidth}
    \includegraphics[height=4.15cm,width=6.8cm]{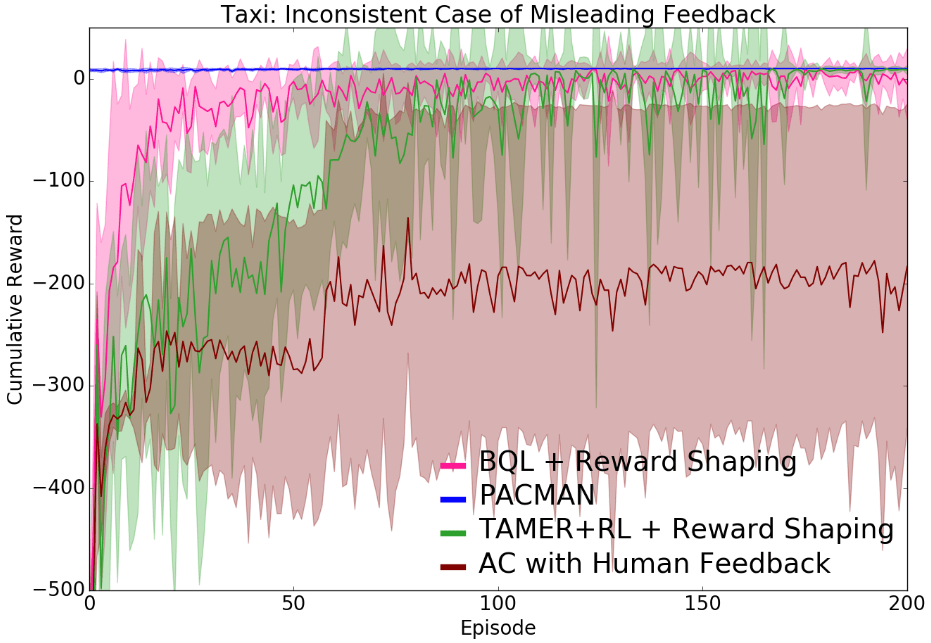}
  \subcaption{Inconsistent Case}
  \label{fig:inconsistent-taxi-mislead}
\end{subfigure}
 \begin{subfigure}{.495\textwidth}
    \includegraphics[height=4.15cm,width=6.8cm]{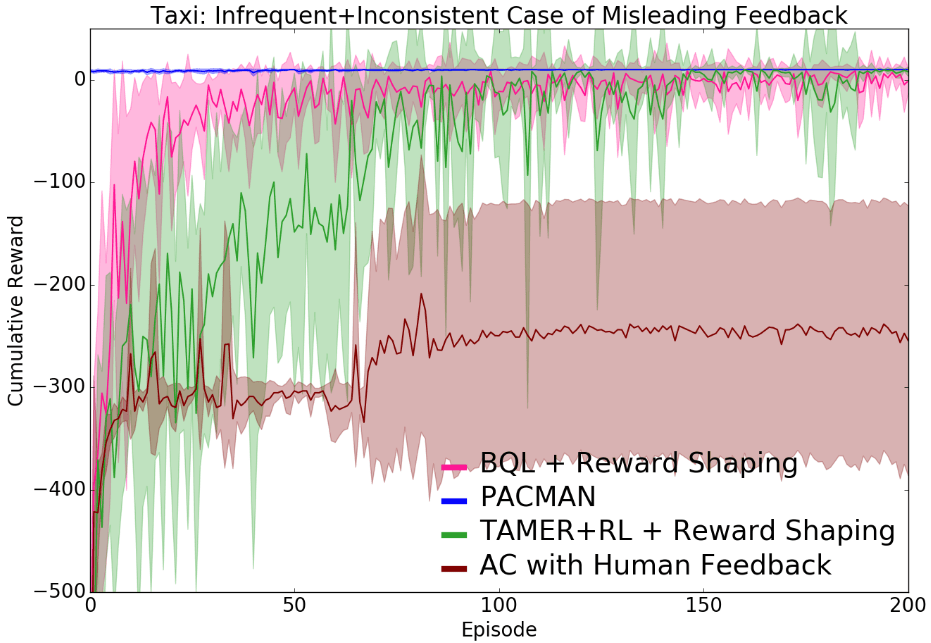}
  \subcaption{Infrequent+Inconsistent Case}
  \label{fig:incons-infreq-taxi-mislead}
\end{subfigure}
\caption{Taxi with Misleading Feedback: Learning Curves}
\label{taximislead}
\end{figure*}


We consider human feedback in the following two scenarios:
\begin{itemize}
\item \textbf{Helpful feedback}. During the rush hour, the passenger can suggest a path that would guide the taxi to detour and avoid the slow traffic, which is shown in Figure~\ref{fig:taxi-help}. The agent should learn a more preferred route from human's feedback.
\item \textbf{Misleading feedback}. Consider a passenger who is not familiar enough with the area and may inaccurately inform the taxi of his location before approaching the passenger (Figure~\ref{fig:taxi-mislead}), which is the wrong action and will mislead the taxi. In this case, the feedback conflicts with symbolic knowledge specified by PACMAN and the agent should learn to ignore such feedback.
\end{itemize}

The results are shown in Figure~\ref{taxihelpful} and Figure~\ref{taximislead}. In the scenario of helpful feedback, the curve of PACMAN has the smallest variance so that it looks like a straight line. But in the case of Infrequent+Inconsistent, there is a big chattering in the initial stage of PACMAN, that's because the symbolic planner is trying some longer plans to do the exploration. In the misleading feedback scenario, the learning speed of the other methods except for PACMAN is quite slow. That's because the human feedback will misguide the agent to perform the improper action that can result in the penalty, and the agent needs a long time to correct its behavior via learning from the environmental reward. But PACMAN keeps unaffected in this case due to the symbolic knowledge that a taxi can pick up the passenger only when it moves to the passenger's location.


\section{Conclusion}
In this paper, we propose the PACMAN framework, which can simultaneously consider prior knowledge, learning from environmental reward and human teaching together and jointly contribute to obtaining the optimal policy.
Experiments show that the PACMAN leads to significant jump-start at early stages of learning, converges faster and with smaller variance, and is robust to inconsistent and infrequent cases even with misleading feedback.

Our future work involves investigating using the PACMAN to perform decision making from high-dimensional sensory input such as pixel images, autonomous driving where the vehicle can learn human's preference on comfort and driving behavior, as well as mobile service robots.

\section{Acknowledgment}
This research was supported in part by the National Science Foundation (NSF) under grants NSF
IIS-1910794. Any opinions, findings, and conclusions or recommendations expressed
in this material are those of the authors and do not necessarily reflect the views of the
NSF.

\nocite{*}
\bibliographystyle{eptcs}
\bibliography{generic}
\end{document}